\documentclass[runningheads]{llncs}

\usepackage{eccv}

\usepackage{eccvabbrv}

\usepackage{amsmath}
\usepackage{graphicx}
\usepackage{booktabs}
\usepackage{enumitem}
\usepackage{pifont}
\usepackage{multirow}
\usepackage{diagbox}
\usepackage[table]{xcolor}

\usepackage[accsupp]{axessibility}  % Improves PDF readability for those with disabilities.

\usepackage{hyperref}
\usepackage{bbding}

\usepackage{orcidlink}

\begin{document}

\title{Robust Trajectory Distillation: Hybrid Reweighting Meets Teacher-Inspired Targets}                                                                                                     % TODO REVIEW: If the paper title is too long for the running head, you can set
\titlerunning{Robust Trajectory Distillation}

%TODO FINAL: Replace with your author list. 
%Include the authors' OCRID for the camera-ready version, if at all possible.
\author{Kaifeng Chen\inst{1}\and
Lechao Cheng\inst{1,2,3,4}\Envelope \and
Jiyang Li\inst{1} \and
Shengeng Tang\inst{1} \and
Fan Zhang\inst{5} \and
Yantao Pan\inst{4} \and
Yaxiong Wang\inst{1} \and
Tuanrui Hui\inst{1} \and
Zhun Zhong\inst{1}
}

\authorrunning{Kaifeng Chen et al.}

\institute{Hefei University of Technology
\and
Jianghuai Advance Technology Center \and
Anhui Provincial Key Laboratory of Humanoid Robots \and
Kaiyang Laboratory,  Chery \and
Zhejiang University of Technology
}

%}                                                                                                                                                                                 %         }
\maketitle

\footnotetext{Corresponding Author: chenglc@hfut.edu.cn}

% 1.摘要需要压缩一下
% 2.增加keywords

\begin{abstract}
% Robust training under noisy labels remains a critical challenge in deep learning due to the risk of confirmation bias and overfitting in iterative correction pipelines. In this work, we propose a novel trajectory-based dataset distillation framework that jointly addresses noise suppression and knowledge preservation without requiring label correction or clean subsets. Our method introduces two complementary components: Selective Guidance Reweighting (SGR) and Teacher-Inspired Auxiliary Targets (TIAT). SGR improves teacher signal quality by integrating global forgetting patterns (via second-split forgetting) with local feature consistency (via KNN-based evaluation), forming a hybrid reweighting mechanism that prioritizes clean supervision. TIAT further enhances the learning capacity by injecting auxiliary guidance derived from intermediate teacher dynamics, ensuring internal consistency while reinforcing informative signals. Together, these strategies enable the distilled dataset to retain cleaner and richer knowledge representations under noisy supervision. The proposed framework is label-preserving, computationally efficient, and broadly applicable. Extensive experiments on benchmark datasets demonstrate consistent performance improvements over state-of-the-art dataset distillation methods across symmetric, asymmetric, and real-world noise scenarios.

Dataset distillation (DD) condenses large corpora into compact, information-rich subsets for efficient training and reuse. However, under noisy supervision, DD risks condensing corrupted associations together with useful signals, degrading robustness. Conventional noisy-label remedies (sample selection, loss weighting, label correction) tightly couple noise estimation with model optimization, often require clean anchors, and can amplify confirmation bias—assumptions that are misaligned with DD’s goal of compact, plug-and-play supervision. We therefore propose a trajectory-based DD framework that jointly suppresses noise and preserves transferable knowledge without relabeling or clean subsets. It comprises two complementary components: Selective Guidance Reweighting (SGR), which fuses global forgetting patterns (second-split forgetting) with local neighborhood consistency into a progressive reweighting scheme that prioritizes clean supervision along the teacher trajectory; and Teacher-Inspired Auxiliary Targets (TIAT), which inject auxiliary residual guidance distilled from intermediate teacher dynamics to reinforce informative signals while remaining internally consistent. Together, SGR and TIAT produce distilled datasets with cleaner and richer representations under noisy supervision. The framework is robust, label-preserving, computationally lightweight, and broadly applicable, yielding consistent gains over state-of-the-art DD baselines across symmetric, asymmetric, and real-world noise.


\end{abstract}    
\section{Introduction}
\label{sec:intro}

The growing scale and complexity of deep models have spurred interest in dataset distillation (DD)—a technique that synthesizes compact, information-rich subsets capable of approximating the training efficacy of large real datasets~\citep{DBLP:journals/corr/abs-1811-10959,cui2022dcbenchdatasetcondensationbenchmark}.

By condensing essential knowledge into a small synthetic set, dataset distillation enables efficient model retraining, rapid adaptation, and even privacy-preserving data sharing. Recent studies~\citep{cheng2024dataset,zhang2024dancedualviewdistributionalignment,guo2024losslessdatasetdistillationdifficultyaligned} further highlight DD’s robustness potential under imperfect supervision, suggesting that distilled datasets can serve as a natural filter that separates informative clean signals from noisy patterns.

However, most existing DD frameworks implicitly assume clean supervision, which severely limits their reliability in real-world noisy-label environments.
Large-scale web-curated datasets often contain substantial annotation noise due to human bias, crowdsourcing inconsistency, and semantic ambiguity~\citep{li2017webvisiondatabasevisuallearning}.
Traditional noisy-label learning techniques—such as sample selection~\citep{malach2017decoupling,liu2020early,zhu2022detecting}, loss weighting~\citep{zhang2018generalized,liu2020peer}, and label correction~\citep{reed2014training,song2019selfie}—attempt to identify and mitigate label corruption during training.
Yet, these methods tightly couple noise estimation and model optimization, forming a self-referential feedback loop where the model acts as both noise detector and predictor.
Such coupling often amplifies confirmation bias, leading to overconfidence in incorrect labels and reduced generalization, while also requiring clean validation anchors or heavy iterative refinement, limiting their scalability.

In this context, dataset distillation under noisy supervision emerges as a compelling alternative paradigm.
Rather than continuously re-estimating noisy labels, DD aims to learn a condensed set of synthetic samples that intrinsically encode cleaner supervision and transferable knowledge.
Nevertheless, two fundamental challenges remain unresolved:

\begin{enumerate}[leftmargin=*]
    \item \textbf{Noise-Agnostic Distillation.} Existing methods such as DANCE~\citep{zhang2024dancedualviewdistributionalignment} and DATM~\citep{guo2024losslessdatasetdistillationdifficultyaligned} lack mechanisms to distinguish corrupted from clean signals during condensation, causing performance degradation under noise (e.g., a 2--3\% accuracy drop on CIFAR-10 at 20\% noise, 50 IPC).
    \item \textbf{Capacity-Constrained Synthesis.} Distilled datasets are typically parameterized as fixed-size image tensors (e.g., 50 IPC), restricting representational capacity and leading to premature information compression before noise--clean disentanglement completes.
\end{enumerate}

These issues motivate our central question:
\emph{How can we improve both the quality and capacity of distilled data under noisy supervision?}
We address this from two complementary perspectives:
\begin{itemize}[leftmargin=*]
    \item \textbf{Challenge \ding{182} (Learning Better):} Improve the fidelity of synthetic data by enhancing teacher-signal reliability and promoting cleaner knowledge retention.
    \item \textbf{Challenge \ding{183} (Learning More):} Increase the amount of high-quality supervision absorbed by the synthetic dataset through richer teacher--student interactions.
\end{itemize}

To this end, we propose two synergistic modules.
\textbf{Selective Guidance Reweighting (SGR)} refines the teacher trajectory with a hybrid KNN--SSFT~\citep{maini2022characterizing,li2022neighborhood} mechanism that integrates local neighborhood consistency with global forgetting trends, yielding cleaner and more trustworthy guidance signals.
\textbf{Teacher-Inspired Auxiliary Targets (TIAT)} supplement the main trajectory with auxiliary residual signals derived from the teacher's internal dynamics—such as intermediate predictions or temporal consistency—thus enriching supervision without introducing conflicting objectives.
Together, SGR and TIAT strengthen both the \emph{quality} and \emph{capacity} of knowledge transfer, enabling robust and scalable dataset distillation under noisy-label conditions.

In summary, we propose a distillation framework specifically tailored for learning under noisy supervision. Conceptually, the framework mirrors a teacher who both (i) refines their expertise to deliver higher-quality instruction and (ii) assigns \textit{homework-like} auxiliary tasks that further reinforce and consolidate the student’s learning. Our contributions are as follows:
\begin{itemize}
    \item We introduce a progressive mechanism that integrates both dynamic and static sample reweighting, fusing complementary teacher-trajectory signals and yielding robust, consistent suppression of label noise.
    \item We propose an auxiliary guidance regularization that enforces consistency with clean trajectories during distillation, amplifying the influence of clean signals throughout training and mitigating error amplification. 
    %\item We propose an auxiliary guidance regularization strategy that ensures clean trajectory consistency during distillation, effectively strengthening the influence of clean samples throughout the process.
    \item  The framework is label-preserving and computationally lightweight: it requires neither relabeling nor costly retraining schedules, making it practical for large-scale, real-world noisy datasets.
     %\item Our framework is label-preserving and computationally efficient, requiring no relabeling or extensive retraining, making it practical for real-world noisy data scenarios.
    \item Extensive experiments across diverse datasets and noise regimes show consistent improvements over strong baselines, demonstrating the method's generality and robustness.
\end{itemize}

\section{Related Works}
\label{sec:related work}

\subsection{Dataset Distillation}
Dataset distillation~\citep{2023datadistillationsurvey, DBLP:journals/corr/abs-1811-10959, cui2022dcbenchdatasetcondensationbenchmark} aims to compress a large dataset into a compact synthetic set while preserving downstream performance. The foundational work~\citep{DBLP:journals/corr/abs-1811-10959} introduced the idea of optimizing synthetic data to match training dynamics observed on real data. Subsequent research has evolved along three major paradigms: \textit{meta-learning}, \textit{parameter matching}, and \textit{distribution matching}. Meta-learning methods~\citep{DBLP:journals/corr/abs-2107-13034, zhou2022datasetdistillationusingneural, loo2023datasetdistillationconvexifiedimplicit} adopt a bi-level optimization framework to generalize across model initializations but suffer from high computational overhead. Parameter matching, including gradient~\citep{zhao2021datasetcondensationgradientmatching} and trajectory matching~\citep{cazenavette2022datasetdistillationmatchingtraining, du2023minimizingaccumulatedtrajectoryerror}, directly aligns model updates between real and synthetic data, with trajectory-based methods achieving state-of-the-art results~\citep{guo2024losslessdatasetdistillationdifficultyaligned}. Variants further explore progressive optimization~\citep{chen2023datadistillationlikevodka}, hybrid data composition~\citep{lee2024selmatcheffectivelyscalingdataset}, and group-wise structures~\citep{he2024multisizedatasetcondensation}. Distribution matching offers an efficient alternative by aligning real and synthetic data distributions in feature space~\citep{zhao2022datasetcondensationdistributionmatching, zhang2024m3ddatasetcondensationminimizing}, class relations~\citep{deng2024exploitingintersampleinterfeaturerelations}, or image-label correlations~\citep{zhang2024dancedualviewdistributionalignment}, without requiring nested optimization. Recent advances even reformulate this as neural feature alignment~\citep{wang2025datasetdistillationneuralcharacteristic}. Beyond dataset-level condensation, model-level knowledge distillation~\citep{zhang2023generalization, wang2024improving}
studies have shown that the formulation of teacher-derived targets is
critical to effective knowledge transfer. For example, regularizing the
direction and norm of student representations toward teacher-derived class
prototypes can substantially improve transfer quality~\cite{wang2024improving}.
Unlike these model-compression settings, our goal is to optimize a compact
synthetic dataset whose induced training dynamics remain reliable under
label noise. However, most methods assume clean supervision and degrade under real-world label noise. To address this limitation, we propose a noise-aware extension of trajectory matching that explicitly models and suppresses corruption during both the distillation and deployment phases.

\subsection{Learning with Noisy Labels}
Real-world datasets often contain corrupted labels due to annotation errors or automated collection. To address this, noisy label learning has developed three major strategies: \textit{sample selection}, \textit{label correction}, and \textit{sample reweighting}. Sample selection methods aim to identify clean data during training, typically using loss-based filtering. A representative work is Decoupling~\citep{malach2017decoupling}, which introduced the small-loss trick. Follow-up works~\citep{jiang2018mentornet, han2018co} incorporate external guidance or co-training, while curriculum-based methods~\citep{lyu2019curriculum, zhou2021robust} and early-learning regularization~\citep{liu2020early} enhance robustness via dynamic filtering or temporal consistency. Some approaches even detect noisy samples prior to training~\citep{zhu2022detecting, wang2018data}. The distinction between noisy and intrinsically difficult samples becomes
particularly challenging under imbalanced class distributions. Fang
et al.~\cite{fang2023separating} addressed this problem by separating
corrupted examples from clean tail-class samples through augmentation
consistency and leave-noise-out regularization. Label correction methods refine labels using model predictions~\citep{reed2014training, zhou2024l2b}, semi-supervised strategies~\citep{li2020dividemix}, or meta-learning and neighborhood consistency~\citep{tu2023learning, li2022neighborhood}. Sample reweighting strategies~\citep{zhang2018generalized, shu2019meta, di2024embedding} instead adjust loss contributions based on sample reliability. Additionally, K-NN-based methods~\citep{bahri2020deep, iscen2022learning} have proven effective for noise detection by evaluating local label consistency in the feature space. Distillation-based robust learning has also been investigated under
imperfect annotations. Reliability-aware sample selection and consistency
regularization can be used to ensure that only trustworthy knowledge is
transferred between models~\cite{fang2023reliable}. Building on this, dataset distillation has recently emerged as a promising alternative. As shown in~\citep{cheng2024dataset}, it addresses limitations such as iterative error amplification and privacy concerns, while offering strong performance under noisy supervision.

\section{Preliminaries}\label{sec:pre}
 \begin{figure*}[t]
  \centering
  \includegraphics[width=1.0\textwidth]{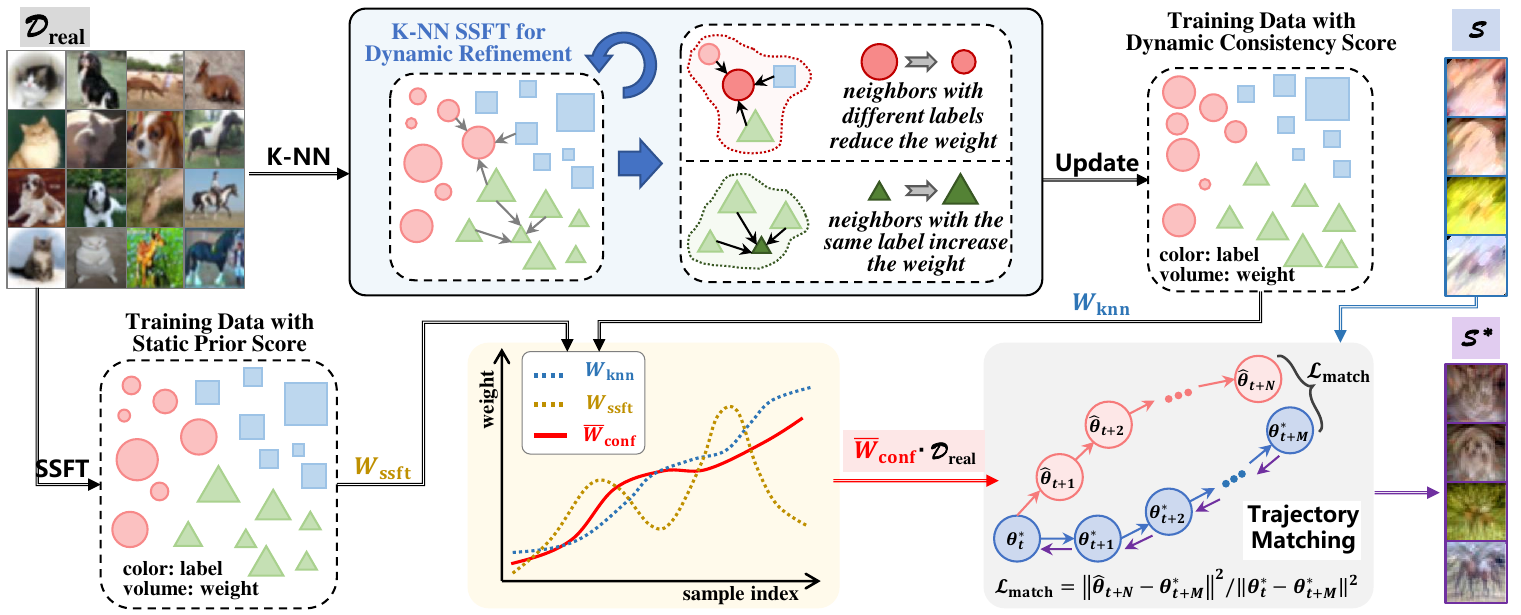}
  \caption{\textbf{The overall pipeline.} Our proposed pipeline includes two main components: (1) during teacher trajectory training, we apply sample-specific weighting adjustments to modulate the influence of each sample based on its estimated reliability; and (2) in the subsequent distillation phase, we leverage a subset of high-confidence samples to impose additional constraints and regularization, thereby enhancing the quality and generalization of the distilled dataset.} 
  \label{fig:pipeline}
  \vspace{-5mm}
\end{figure*}
\subsection{Dataset Distillation with Noisy Labels}
Let $\mathcal{D}_\text{real} = \{(x_i, \tilde{y}_i)\}_{i=1}^{N}$ denote a real-world training dataset, where $\tilde{y}_i$ is the observed (and potentially noisy) label for input $x_i$. We assume the existence of label noise such that $\tilde{y}_i \neq y_i^*$ for some $i$, where $y_i^*$ is the clean but unobserved ground-truth label. In contrast, the test set $\mathcal{D}_\text{test}$ is assumed to be entirely clean and representative of the true data distribution, and is used to evaluate generalization performance. Our objective is to synthesize a compact dataset $\mathcal{S}$, with $|\mathcal{S}| \ll |\mathcal{D}_\text{real}|$, such that a model trained solely on $\mathcal{S}$ achieves lower generalization error on $\mathcal{D}_\text{test}$ than one trained on the full noisy dataset $\mathcal{D}_\text{real}$. Formally, the distillation task can be formulated as the following optimization problem:
\setlength{\abovedisplayskip}{3pt}
\setlength{\abovedisplayshortskip}{3pt}
\setlength{\belowdisplayskip}{3pt}
\setlength{\belowdisplayshortskip}{3pt}
\begin{equation} \label{eq:general-obj}
    \mathcal{S}^* = \arg\min_{\mathcal{S}} \, \mathcal{L}(\mathcal{S}, \mathcal{D}_\text{real}),
\end{equation}
where $\mathcal{L}$ is a general objective function. 
Following prior trajectory-matching approaches~\citep{guo2024losslessdatasetdistillationdifficultyaligned}, we instantiate $\mathcal{L}$ as a loss that aligns the learning dynamics of a student trained on $\mathcal{S}$ with those of a teacher trained on $\mathcal{D}_\text{real}$.
%In our setting, it is instantiated as a trajectory matching loss~\citep{guo2024losslessdatasetdistillationdifficultyaligned} that aligns the student’s learning dynamics (trained on $\mathcal{S}$) with those of a teacher model trained on $\mathcal{D}_\text{real}$.
Trajectory matching-based dataset distillation methods typically adopt a bilevel optimization scheme comprising an \textit{inner loop} and an \textit{outer loop}. The inner loop simulates the training dynamics of a student model on the current synthetic dataset $\mathcal{S}$, while the outer loop updates $\mathcal{S}$ such that the student’s optimization trajectory closely matches that of a teacher trained on the real dataset $\mathcal{D}_\text{real}$.

Specifically, a teacher model is first trained on $\mathcal{D}_\text{real}$ to produce a reference trajectory $\tau^* = \{\theta^*_t\}_{t=1}^{T}$, where $\theta^*_t$ denotes the model parameters at iteration $t$. Meanwhile, the synthetic dataset $\mathcal{S}$ is initialized—either by sampling from Gaussian noise or selecting real samples—with corresponding soft or hard labels.

In the \textit{inner loop}, we simulate student learning dynamics by iteratively updating parameters using $\mathcal{S}$. Given a starting point $\hat{\theta}_t = \theta^*_t$, the student parameters are updated over $N$ steps according to:
% \begin{equation}
%     \hat{\theta}_{t+n+1} = \hat{\theta}_{t+n} - \alpha \nabla_{\hat{\theta}_{t+n}} \ell(\mathcal{A}(\mathcal{S}); \hat{\theta}_{t+n}),
% \end{equation}
\begin{equation}
    \hat{\theta}_{t+n+1}
    = \hat{\theta}_{t+n} - \alpha \nabla_{\hat{\theta}_{t+n}}
      \ell(\mathcal{A}_n(\mathcal{S}); \hat{\theta}_{t+n}),
    \quad n = 0,\dots,N-1,
\end{equation}
where $\ell$ denotes the task loss (e.g., cross-entropy), $\mathcal{A}(\mathcal{S})$ denotes a mini-batch drawn from $\mathcal{S}$ with optional data augmentation, and $\alpha$ is the learning rate. This simulates the student trajectory $\hat{\tau} = \{\hat{\theta}_{t+n}\}_{n=0}^N$ induced by $\mathcal{S}$.

In the \textit{outer loop}, we optimize the synthetic dataset $\mathcal{S}$ such that the student’s final parameters align with the teacher’s future trajectory. Let $\mathcal{T} = \{t_1, t_2, \dots, t_{max}\}$ be a set of anchor steps. At each $t \in \mathcal{T}$, the teacher parameters after $M$ additional steps are denoted as $\theta^*_{t+M}$. 
In our method, $\mathcal{L}$ in Eq.~\ref{eq:general-obj} is instantiated as the following normalized trajectory alignment loss:
% \begin{equation} \label{eq:traj-loss}
%     \mathcal{L}_\text{traj}(\mathcal{S}, \mathcal{D}_\text{real}) =
%     \frac{\left\| \hat{\theta}_{t+N} - \theta^*_{t+M} \right\|_2^2}{\left\| \theta^*_{t} - \theta^*_{t+M} \right\|_2^2}.
% \end{equation}
\begin{equation} \label{eq:traj-loss}
    \mathcal{L}_\text{traj}
    =  \frac{\left\| \hat{\theta}_{t+N} - \theta^*_{t+M} \right\|_2^2}
           {\left\| \theta^*_t - \theta^*_{t+M} \right\|_2^2}.
\end{equation}

The normalization term $\left\| \theta^*_{t} - \theta^*_{t+M} \right\|_2^2$ calibrates the alignment error relative to the teacher's own parameter change, making the loss scale-invariant across steps.
% The trajectory matching loss is defined as:
% \begin{equation}
%     \mathcal{L}_\text{traj}(\mathcal{S}, \mathcal{D}_\text{real}) =
%      \frac{\left\| \hat{\theta}_{t+N} - \theta^*_{t+M} \right\|_2^2}{\left\| \theta^*_{t} - \theta^*_{t+M} \right\|_2^2}.
% \end{equation}
The numerator measures how closely the student, trained on $\mathcal{S}$, approximates the teacher’s future parameters, while the denominator normalizes for the teacher’s update magnitude to ensure scale invariance. This outer-loop objective is used to update $\mathcal{S}$ via gradient-based optimization, enabling the synthetic data to induce faithful learning dynamics that reflect those observed on real data.

\section{Methodology}
\label{section:Methods}

\subsection{Overview}
Figure~\ref{fig:pipeline} illustrates the overall workflow of our proposed framework for noisy dataset distillation, which aims to construct a compact synthetic dataset $\mathcal{S}$ from a noisy real-world dataset $\mathcal{D}_\text{real} = \{(x_i, \tilde{y}_i)\}_{i=1}^{N}$. The core idea is to train the student such that it replicates the learning dynamics of a teacher model trained on $\mathcal{D}_\text{real}$, while suppressing the adverse effects of label noise.
To this end, our framework follows a three-stage pipeline. First, we train a teacher model on $\mathcal{D}_\text{real}$ to obtain a reference parameter trajectory $\tau^* = {\theta_t^*}$; during this stage, we apply \textit{Selective Guidance Reweighting (SGR)} (cf. Sec.~\ref{section:SGR}) to assign sample-specific weights based on reliability estimates derived from second-split forgetting and KNN-based local consistency, thereby producing a cleaner supervision signal. Second, we initialize a synthetic dataset $\mathcal{S}$ with soft or hard labels and simulate the inner-loop training dynamics of a student model on $\mathcal{S}$. In this stage, we further incorporate \textit{Teacher-Inspired Auxiliary Targets (TIAT)} (cf. Sec.~\ref{section:TIAT}) —a set of auxiliary consistency signals extracted from intermediate teacher states over high-confidence samples — to boost supervision beyond trajectory alignment. These high-confidence real samples are used only to refine teacher checkpoints and construct auxiliary targets, they are not paired with, indexed against, or directly mapped to individual synthetic samples. Finally, in the outer loop, we optimize $\mathcal{S}$ via a normalized trajectory matching loss computed across multiple anchor steps, aligning the student’s learned trajectory with that of the teacher. This bilevel optimization process jointly improves the quality of supervision (via SGR) and the effectiveness of trajectory alignment (via TIAT), enabling the distilled dataset to retain cleaner, more generalizable knowledge even under noisy conditions.

\subsection{Selective Guidance Reweighting}
\label{section:SGR}

We propose \textbf{Selective Guidance Reweighting (SGR)} to control the fidelity of signals used for synthetic data optimization. SGR introduces a hybrid reliability estimator for each training sample $i$, composed of:

\begin{itemize}[leftmargin=*, topsep=0pt]
    \item a \textbf{dynamic consistency score} $W_{\text{knn}}^{(i)}$ based on feature-space neighborhood agreement.
    \item a \textbf{static prior score} $W_{\text{ssft}}^{(i)}$ derived from sample forgetting behavior;
\end{itemize}

These scores are combined into a unified weight $W^{(i)}$ via time-dependent convex interpolation.

\paragraph{Dynamic Estimation via KNN.}
Given the predicted label distributions $\hat{y}_j$ of the $K$ nearest neighbors of the observed sample $i$ in the feature space, we define its KNN consistency score as:
\begin{equation}
W_{\text{knn}}^{(i)} = 1 - \frac{1}{K} \sum_{j=1}^{K} \mathrm{JS}(\tilde{y}_i, \hat{y}_j),
\end{equation}
where $\tilde{y}_i$ is viewed as a one-hot distribution and $\mathrm{JS}(\cdot,\cdot)$ denotes the Jensen--Shannon divergence. This score ranges in $[0, 1]$, with higher values indicating better local agreement.

\paragraph{Static Prior via SSFT.}
Inspired by~\citep{maini2022characterizing}, we introduce a forgetting-based difficulty score to quantify the reliability of each training sample. For each sample $i$, we record:
\begin{itemize}[leftmargin=*, topsep=0pt]
    \item $t^{(i)}_{\text{learn}}$: the earliest epoch when the sample is first correctly classified;
    \item $t^{(i)}_{\text{forget}}$: the earliest epoch after which it is misclassified again.
\end{itemize}

These timestamps reflect the memorization and retention behavior of a sample during training. To assess sample-level difficulty, we define the SSFT score as:
\begin{equation}
s^{(i)} = \lambda \cdot \frac{t^{(i)}_{\text{learn}}}{\max_j t^{(j)}_{\text{learn}}} 
+ (1 - \lambda) \cdot \left(1 - \frac{t^{(i)}_{\text{forget}}}{\max_j t^{(j)}_{\text{forget}}} \right),
\label{eq:ssft-score}
\end{equation}
where $\lambda \in [0, 1]$ balances the contributions of learnability and forgettability. A high $s^{(i)}$ implies that sample $i$ was learned late and forgotten early, thus more likely to be noisy or difficult. We convert this difficulty score into a static reliability weight:
\begin{equation}
W_{\text{ssft}}^{(i)} = 1 - s^{(i)},
\end{equation}
such that clean and stable samples are assigned higher weights at the beginning of training. 
This reflects the global memorization difficulty of sample $i$: higher $W_{\text{ssft}}^{(i)}$ values favor clean and stable samples, while lower values tend to down-weight potentially noisy or hard-to-learn examples.
%This reflects the global memorization difficulty of sample $i$, with higher scores favoring clean samples with high probabilities.

\paragraph{Hybrid Weighting via Curriculum Fusion.}
To balance global priors and evolving local consistency, we define a convex combination:
\begin{equation}
W_t^{(i)} = (1 - \alpha_t) \cdot W_{\text{ssft}}^{(i)} + \alpha_t \cdot W_{\text{knn}}^{(i)},
\end{equation}
where $\alpha_t \in [0, 1]$ is a time-dependent blending coefficient that increases linearly over training epochs. Specifically, we set:
\begin{equation}
\alpha_t = \min\left( \frac{t}{T_{\text{warmup}}} \cdot \alpha_{\max}, \alpha_{\max} \right),
\end{equation}
where $T_{\text{warmup}}$ is a predefined transition period (e.g., 20\% of training). Optionally, when training multiple teacher trajectories indexed by $p \in \{1, \dots, P\}$, we set $\alpha_{\max}^{(p)} = \frac{p-1}{P}$ to diversify the static-dynamic balance across teachers (more details on multi-trajectory aggregation are given in Eq.~\ref{eq:pca}).
\paragraph{Remark.}
The timestep weight $W_t^{(i)}$ is used to modulate the contribution of sample $i$ when updating the teacher model parameters or computing guidance for $\mathcal{S}$ in outer-loop optimization. This strategy enables the distillation process to prioritize clean, informative signals throughout training.

\subsection{Teacher-Inspired Auxiliary Targets (TIAT)}
\label{section:TIAT}
Although the Selective Guidance Reweighting (SGR) module effectively suppresses noisy signals along the teacher trajectory, its role is essentially confined to improving the quality of teacher-side guidance, and it lacks an explicit mechanism to guarantee that the distilled dataset $\mathcal{S}$ remains consistently aligned with clean supervision in the representation space. More concretely, SGR amplifies cleaner and more consistent teacher signals, thereby enhancing the reliability of the teacher trajectory itself, but it does not directly regulate how the student model absorbs and responds to these reweighted signals during trajectory alignment and parameter updates. As a consequence, even after reweighting, the student optimization can still overfit residual noise or locally unstable regions in the teacher trajectory, manifesting as decision boundary shifts and limited generalization, and thus failing to fully realize the theoretical benefits of noise suppression.

To address this issue, we propose \textbf{Teacher-Inspired Auxiliary Targets (TIAT)}, 
a complementary mechanism that operates on the student side. 
TIAT injects clean-aware regularization into the distillation process by identifying 
a reliable subset and modeling trajectory-level uncertainty, while further incorporating 
auxiliary supervision derived from intermediate, high-confidence teacher states. 
This design effectively regularizes the student's optimization, promoting more stable 
and consistent learning dynamics. By jointly optimizing trajectory alignment 
(Eq.~\ref{eq:lossMatch}) and uncertainty-aware auxiliary objectives 
(Eq.~\ref{eq:lossAux}--Eq.~\ref{eq:lossTotal}), TIAT bridges the gap between 
teacher-signal refinement and student optimization stability, ensuring that the 
synthetic dataset learns predominantly from the most trustworthy stages of the 
teacher's knowledge evolution.

\paragraph{Probabilistic Confidence Aggregation.}
Let $W_{p}^{(i)}$ denote the final-stage reliability weight assigned to sample $i$ by the $p$-th teacher trajectory, trained with static-dynamic blending coefficient $\alpha^{(p)}_{\max}$. To estimate the sample's overall confidence, we define:
\begin{equation}\label{eq:pca}
\bar{W}_{\text{conf}}^{(i)} := \mathbb{E}_{p \sim \mathcal{U}(\mathcal{P})} \left[ W_{p}^{(i)} \right] 
\approx \frac{1}{P} \sum_{p=1}^{P} W_{p}^{(i)},
\end{equation}
where $\mathcal{P} = \{1, \dots, P\}$ is the index set of all trajectories. This aggregated score summarizes the expected reliability of sample $i$ across all teacher views.

We then compute the probabilistic trajectory alignment loss $\mathcal{L}_{\text{match}}$ that encourages student updates to follow the aggregated teacher behavior:
\begin{equation}\label{eq:lossMatch}
\mathcal{L}_{\text{match}} = \frac{\|\hat{\theta}_{t+N}^{(i)} - \theta^{*}_{t+M}\|_2^2}{\|\theta^{*}_{t+M} - \theta^{*}_{t}\|_2^2},
\end{equation}
where $\hat{\theta}_{t+N}^{(i)}$ denotes the student parameters after being updated on synthetic sample $i$ for $N$ optimization steps starting from the teacher state at iteration $t$.

\paragraph{Uncertainty-Aware Auxiliary Regularization.}
To quantify confidence variance across trajectories, we compute:
\begin{equation}\label{eq:aar}
\operatorname{Var}_{p}(W_{p}^{(i)}) = \mathbb{E}_{p \sim \mathcal{U}(\mathcal{P})} \left[ \left( W_{p}^{(i)} - \bar{W}_{\text{conf}}^{(i)} \right)^2 \right].
\end{equation}
This variance reflects the stability of confidence scores assigned to sample $i$; low-variance samples are deemed more consistently reliable. 

We then define an approximately reliable subset $\mathcal{D}_{\text{sub}}$ using two complementary criteria: $W_{\text{ssft}}^{(i)} \geq \delta_{\text{sup}}$ and $\operatorname{Var}_{p}(W_{p}^{(i)}) \leq \sigma_{\text{inf}}$, where $\delta_{\text{sup}}$ and $\sigma_{\text{inf}}$ are fixed thresholds. This set captures samples that are both statistically confident and dynamically stable, forming the basis of our auxiliary regularization. Based on $\mathcal{D}_{\text{sub}}$, we fine-tune $\theta^*_t$ to produce a cleaner teacher checkpoint $\theta^{\text{ft}}_{t+M}$ and define the auxiliary loss as:
\begin{equation}\label{eq:lossAux}
\mathcal{L}_{\text{aux}} = \frac{\|\hat{\theta}_{t+N}^{(i)} - \theta^{\text{ft}}_{t+M}\|_2^2}{\|\theta^{\text{ft}}_{t+M} - \theta^{*}_{t}\|_2^2}.
\end{equation}

To obtain the refined teacher checkpoint $\theta^{\text{ft}}_{t+M}$, we perform a short fine-tuning of the teacher model using only the reliable subset $\mathcal{D}_{\text{sub}}$. This fine-tuning process retains the same learning protocol as the original teacher but is restricted to several additional epochs on high-confidence samples, effectively filtering out the influence of noisy labels. The resulting $\theta^{\text{ft}}_{t+M}$ thus represents a noise-suppressed continuation of the teacher trajectory.

To consolidate the complementary contributions of the original and refined trajectories, 
we formulate a unified objective function as follows:
\begin{equation}\label{eq:lossTotal}
\mathcal{L}_{\text{total}} = (1 - \beta) \cdot \mathcal{L}_{\text{match}} + \beta \cdot \mathcal{L}_{\text{aux}}, \quad \beta \in [0,1],
\end{equation}
%where $\beta$ balances the influence of the original and clean-adjusted trajectories.
where $\beta$ balances the influence of the original teacher trajectory and the clean-adjusted auxiliary trajectory.

\paragraph{Discussion.}
TIAT introduces an uncertainty-aware auxiliary supervision signal into dataset distillation, driven by both trajectory consensus and variance-aware selection. By formulating confidence as a probabilistic mean and explicitly modeling uncertainty, the method enables soft guidance that avoids hard filtering while remaining computationally efficient. This module complements SGR by enforcing trajectory-level regularization from the student side, forming a closed-loop distillation pipeline that is robust to noisy supervision.

\section{Experiments}

\subsection{Experimental Setup}

We evaluate our method on two widely used image classification datasets, CIFAR-10/100~\citep{Krizhevsky2009LearningML} and Tiny-ImageNet~\citep{le2015tiny}, under noisy-label settings. 
Comprehensive experiments are conducted across different noise regimes and images-per-class (IPC) configurations to assess the robustness and scalability of our approach in the context of noisy label learning (LNL) and dataset distillation (DD).

\noindent\textbf{Noise Settings.}
Following prior work~\citep{song2022learning,englesson2024robust,iscen2022learning}, we consider both synthetic and real-world label noise:

\begin{itemize}[leftmargin=*]

\item \textbf{Symmetric Noise.}
Labels are uniformly corrupted across all non-target classes. 
Given a noise rate $\eta$ and $C$ classes, the label transition is defined as
\[
P(y^{\text{noisy}} = c' \mid y^{\text{clean}} = c) =
\begin{cases}
1 - \eta, & c' = c,\\[2pt]
\frac{\eta}{C - 1}, & c' \neq c,
\end{cases}
\]
where $c, c' \in \{1,\dots,C\}$.
We set $\eta \in \{20\%, 40\%\}$ in our experiments.

\item \textbf{Asymmetric Noise.}
Labels are flipped according to class-dependent transition rules, which reflect realistic confusions (e.g., \emph{cat} $\leftrightarrow$ \emph{dog}, \emph{truck} $\rightarrow$ \emph{automobile}).
We adopt the standard asymmetric transition matrices used in prior LNL studies~\citep{song2022learning,iscen2022learning}.

\item \textbf{Real-World Noise.}
We additionally evaluate on CIFAR-N~\citep{wei2021learning}, a human-annotated variant of CIFAR-10/100 that exhibits natural labeling inconsistencies.
Its multi-annotator design captures human ambiguity patterns, providing a realistic benchmark for noisy labels.
\end{itemize}

\noindent\textbf{Implementation Details.}
We compare our method against the following baselines:
(i) a \emph{noisy baseline}, which trains directly on the corrupted dataset without any distillation; (ii) a \emph{random subset selection} baseline, which constructs condensed datasets by uniformly sampling the same number of images per class from the noisy set under the given IPC budget;
 and (iii) state-of-the-art dataset distillation approaches, including DATM~\citep{guo2024losslessdatasetdistillationdifficultyaligned}, DANCE~\citep{zhang2024dancedualviewdistributionalignment}, RDED~\citep{sun2024diversity}, and RCIG~\citep{loo2023datasetdistillationconvexifiedimplicit}. 
To ensure a fair comparison, we follow the key experimental settings in prior work, particularly DATM.
Specifically, we adopt a three-layer ConvNet backbone for CIFAR-10/100 and a four-layer ConvNet for Tiny-ImageNet in all methods.
For evaluation, we report the mean and standard deviation of test accuracy over $5$ independently initialized networks trained on the distilled dataset $\mathcal{S}$. 
Unless otherwise specified, we follow DATM to use $P=100$ teacher trajectories. For SGR, we set the KNN neighborhood size to $K=10$, the SSFT balance coefficient to $\lambda=0.6$, and the warm-up length to $T_{warmup}=60$. The SSFT timestamps are generated once before distillation and reused across experiments, introducing only one-time preprocessing overhead. For TIAT, we use a 60\% high-confidence subset ratio and set $\beta=0.1$, as validated in Table~\ref{tab:high_confidence_ratio}.

\subsection{Benchmarking Dataset Distillation Results on Noisy Dataset}

As shown in Table~\ref{tab:vsSota CIFAR-10 (a)symmetric}, across all four noise settings (symmetric/asymmetric at 20\% and 40\%), 
our method achieves the strongest overall robustness across noise settings, especially under higher noise rates and larger IPC budgets, while DANCE remains competitive in several low-IPC cases.
% our method consistently achieves higher test accuracy than four state-of-the-art baselines---RCIG, RDED, DANCE, and DATM. 
The advantage is particularly pronounced in high-IPC regimes and under severe noise (40\%), where competing methods degrade rapidly, while our method maintains both strong accuracy and smooth performance trends. 
Even under the challenging asymmetric noise setting, where existing dataset distillation methods typically incur substantial performance drops, our approach remains consistently superior at both 20\% and 40\% noise rates. 
These observations indicate that our clean-aware sample reweighting and trajectory-informed distillation substantially improve robustness to label corruption, especially when the training data is simultaneously scarce and noisy.

% \begin{figure}[t]
%   \centering
%   \includegraphics[width=\textwidth]{fig/CmpSota/a,symVsSota_old.pdf}
%   \caption{\centering Test accuracy (\%) on CIFAR-10 under (a)symmetric noise (20\% and 40\%).} 
%   \label{fig:vsSota CIFAR-10 (a)symmetric}
% \end{figure}

\begin{table}[tb]
  \renewcommand{\arraystretch}{1.3}
  \caption{\centering Test accuracy (\%) on CIFAR-10 under (a)symmetric noise (20\% and 40\%).} 
  \centering
  \vspace{-2mm}
  \setlength{\tabcolsep}{3pt}
  \resizebox{\textwidth}{!}{
  \large

  \begin{tabular}{c|l|cccc|cccc}
\hline
\multirow{2}{*}{\textbf{Noise Type}} 
& \multirow{2}{*}{\diagbox[width=7em]{\textbf{Method}}{\textbf{Noise Ratio}}}
& \multicolumn{4}{c|}{\textbf{20\%}} 
& \multicolumn{4}{c}{\textbf{40\%}} \\
& 
& IPC=10 & 50 & 500 & 1000 
& 10 & 50 & 500 & 1000 \\
\hline

\multirow{6}{*}{\textbf{Symmetric}}
& \textbf{Full Dataset}
& \multicolumn{4}{c|}{76.1$\pm$0.2}
& \multicolumn{4}{c}{66.2$\pm$0.5} \\

& \textbf{Subset}
& 25.1$\pm$0.3 & 38.8$\pm$0.5 & 53.2$\pm$0.7 & 57.1$\pm$0.6
& 15.3$\pm$0.5 & 27.6$\pm$0.5 & 38.9$\pm$0.7 & 43.1$\pm$1.0 \\

& \textbf{RCIG}~\citep{loo2023datasetdistillationconvexifiedimplicit}
& 67.1$\pm$0.3 & 71.5$\pm$0.3 & 72.0$\pm$0.2 & -
& 65.6$\pm$0.3 & 67.8$\pm$0.5 & 68.9$\pm$0.3 & - \\

& \textbf{RDED}~\citep{sun2024diversity}
& 52.3$\pm$0.8 & 67.3$\pm$0.9 & 74.9$\pm$0.4 & -
& 52.2$\pm$0.8 & 62.3$\pm$0.3 & 65.9$\pm$0.4 & - \\

& \textbf{DANCE}~\citep{zhang2024dancedualviewdistributionalignment}
& \textbf{68.7$\pm$0.4} & 71.9$\pm$0.3 & 73.0$\pm$0.2 & -
& \textbf{65.8$\pm$0.2} & 67.6$\pm$0.2 & 68.4$\pm$0.3 & - \\

& \textbf{DATM}~\citep{guo2024losslessdatasetdistillationdifficultyaligned}
& 62.7$\pm$0.6 & 71.7$\pm$0.3 & 80.0$\pm$0.4 & 81.9$\pm$0.1
& 58.2$\pm$0.2 & 67.2$\pm$0.3 & 74.3$\pm$0.3 & 76.7$\pm$0.2 \\

& \textbf{Ours}
& 64.5$\pm$0.3 & \textbf{73.5$\pm$0.4} & \textbf{81.5$\pm$0.2} & \textbf{83.3$\pm$0.3}
& 64.8$\pm$0.5 & \textbf{71.6$\pm$0.3} & \textbf{80.0$\pm$0.2} & \textbf{81.4$\pm$0.1} \\

\hline

\multirow{6}{*}{\textbf{Asymmetric}}
& \textbf{Full Dataset}
& \multicolumn{4}{c|}{80.2$\pm$0.2}
& \multicolumn{4}{c}{71.1$\pm$0.4} \\

& \textbf{Subset}
& 30.0$\pm$0.4 & 44.1$\pm$0.2 & 63.2$\pm$0.9 & 68.8$\pm$0.3
& 25.2$\pm$0.3 & 40.0$\pm$1.1 & 59.8$\pm$0.6 & 64.8$\pm$1.1 \\

& \textbf{RCIG}~\citep{loo2023datasetdistillationconvexifiedimplicit}
& 64.6$\pm$0.3 & 69.9$\pm$0.3 & 71.0$\pm$0.3 & -
& 58.9$\pm$0.6 & 63.1$\pm$0.3 & 64.2$\pm$0.2 & - \\

& \textbf{RDED}~\citep{sun2024diversity}
& 46.1$\pm$1.6 & 64.8$\pm$0.5 & 75.4$\pm$0.5 & -
& 43.6$\pm$1.1 & 60.0$\pm$1.3 & 69.3$\pm$1.2 & - \\

& \textbf{DANCE}~\citep{zhang2024dancedualviewdistributionalignment}
& \textbf{69.2$\pm$0.3} & 73.4$\pm$0.1 & 73.6$\pm$0.4 & -
& \textbf{66.1$\pm$0.2} & \textbf{69.5$\pm$0.1} & 69.6$\pm$0.2 & - \\

& \textbf{DATM}~\citep{guo2024losslessdatasetdistillationdifficultyaligned}
& 61.9$\pm$0.4 & 71.3$\pm$0.4 & 77.7$\pm$0.3 & 80.4$\pm$0.3
& 55.5$\pm$0.3 & 64.0$\pm$0.3 & 69.7$\pm$0.5 & 71.4$\pm$0.3 \\

& \textbf{Ours}
& 63.7$\pm$0.8 & \textbf{73.6$\pm$0.3} & \textbf{79.8$\pm$0.3} & \textbf{82.1$\pm$0.3}
& 58.7$\pm$0.5 & 67.0$\pm$0.5 & \textbf{73.2$\pm$0.4} & \textbf{75.0$\pm$0.4} \\

\hline
\end{tabular}
}
\vspace{-5mm}
\label{tab:vsSota CIFAR-10 (a)symmetric}
\end{table}

On real-world noisy labels (CIFAR-N, Table~\ref{tab:vsSota CIFAR-10N}), most methods overfit as IPC increases, with accuracy saturating or even degrading. In contrast, trajectory-matching approaches (DATM and ours) remain robust and benefit more from larger IPC, indicating a stronger ability to extract clean supervision. As noise grows from Aggre (9\%) to Worst (40\%), all methods degrade, but our approach consistently leads, with the margin widening at higher IPC and exceeding the best baseline by over 3\% from IPC~50 onward. These results demonstrate the strong noise resilience and compression robustness of our framework across diverse noise levels and data scales.

\begin{table}
  \setlength{\tabcolsep}{3pt}      
  \renewcommand{\arraystretch}{0.95} 
  \vspace{-5mm}
  \caption{ Test accuracy (\%) on \textbf{CIFAR-10N}.}
  \vspace{-2mm}
  \centering
  \resizebox{0.85\textwidth}{!}{
  \small
  \begin{tabular}{c|l|cccc|c}
\toprule
\multirow{2}{*}{\textbf{Noise Type}} 
& \multirow{2}{*}{\diagbox[width=7em]{\textbf{Method}}{\textbf{IPC}}}
& \multicolumn{4}{c|}{} 
& \textbf{Full} \\
& & 10 & 50 & 500 & 1000 & \textbf{Dataset} \\
\midrule

\multirow{6}{*}{\shortstack{\textbf{Aggre}\\(9.03\%)}} 

& \textbf{Subset}
& 31.5$\pm$0.6 & 43.4$\pm$0.3 & 63.7$\pm$0.9 & 69.7$\pm$0.5 &  \\

& \textbf{RCIG}~\citep{loo2023datasetdistillationconvexifiedimplicit}
& 66.5$\pm$0.2 & 72.9$\pm$0.3 & - & - &  \\

& \textbf{RDED}~\citep{sun2024diversity}
&  48.4$\pm$1.1 & 65.4$\pm$1.9 & 77.2$\pm$0.4 & - &  \\

& \textbf{DANCE}~\citep{zhang2024dancedualviewdistributionalignment}
& \textbf{69.7$\pm$0.4} & \textbf{74.1$\pm$0.4} & 74.6$\pm$0.3 & - & 82.5$\pm$0.4 \\

& \textbf{DATM}~\citep{guo2024losslessdatasetdistillationdifficultyaligned}
& 63.7$\pm$0.5 & 73.2$\pm$0.2 & 80.4$\pm$0.2 & 83.0$\pm$0.3 &  \\

& \textbf{Ours}
& 65.2$\pm$0.4 & 73.5$\pm$0.3 & \textbf{80.8$\pm$0.1} & \textbf{83.4$\pm$0.2} &  \\

\midrule

\multirow{6}{*}{\shortstack{\textbf{Rand1}\\(18\%)}} 

& \textbf{Subset}
& 28.9$\pm$0.8 & 40.9$\pm$0.5 & 60.0$\pm$0.6 & 63.0$\pm$0.5 &  \\

& \textbf{RCIG}~\citep{loo2023datasetdistillationconvexifiedimplicit}
& 66.9$\pm$0.2 & 72.4$\pm$0.2 & - & - &  \\

& \textbf{RDED}~\citep{sun2024diversity}
& 49.2$\pm$1.5 & 67.0$\pm$0.8 & 77.2$\pm$1.0 & - &  \\

& \textbf{DANCE}~\citep{zhang2024dancedualviewdistributionalignment}
& \textbf{69.8$\pm$0.3} & 73.3$\pm$0.4 & 73.8$\pm$0.3 & - & 79.4$\pm$0.1 \\

& \textbf{DATM}~\citep{guo2024losslessdatasetdistillationdifficultyaligned}
& 63.3$\pm$0.5 & 72.5$\pm$0.3 & 79.4$\pm$0.3 & 81.8$\pm$0.3 &  \\

& \textbf{Ours}
& 64.5$\pm$0.5 & \textbf{73.6$\pm$0.3} & \textbf{80.6$\pm$0.3} & \textbf{82.9$\pm$0.1} &  \\

\midrule

\multirow{6}{*}{\shortstack{\textbf{Worst}\\(40.21\%)}} 

& \textbf{Subset}
& 22.2$\pm$0.5 & 35.8$\pm$0.4 & 46.2$\pm$0.4 & 50.0$\pm$1.1 &  \\

& \textbf{RCIG}~\citep{loo2023datasetdistillationconvexifiedimplicit}
& 65.3$\pm$0.3 & 67.7$\pm$0.4 & - & - &  \\

& \textbf{RDED}~\citep{sun2024diversity}
& 50.6$\pm$1.3 & 65.2$\pm$1.4 & 72.5$\pm$0.8 & - &  \\

& \textbf{DANCE}~\citep{zhang2024dancedualviewdistributionalignment}
& \textbf{65.8$\pm$0.3} & 68.5$\pm$0.2 & 68.8$\pm$0.1 & - & 67.1$\pm$0.2 \\

& \textbf{DATM}~\citep{guo2024losslessdatasetdistillationdifficultyaligned}
& 58.8$\pm$0.6 & 67.8$\pm$0.3 & 73.2$\pm$0.4 & 75.3$\pm$0.5 &  \\

& \textbf{Ours}
& 62.4$\pm$0.6 & \textbf{70.8$\pm$0.5} & \textbf{77.1$\pm$0.4} & \textbf{79.2$\pm$0.4} &  \\

\bottomrule
\end{tabular}
}
\vspace{-6mm}
\label{tab:vsSota CIFAR-10N}
\end{table}

Table~\ref{tab:vsSota CIFAR-100} and Table~\ref{tab:vsSota CIFAR-100N} further compares our method with recent state-of-the-art approaches on \textbf{CIFAR-100N} and \textbf{CIFAR-100} under symmetric and asymmetric noise at 20\% and 40\% corruption ratios. 
Across all IPC levels, our method attains the best accuracy in nearly all settings. 
Notably, under 40\% symmetric noise, we outperform prior methods by substantial margins, achieving \textbf{52.8\%} at 100 IPC and \textbf{43.2\%} at 10 IPC. 
Even in the challenging \textbf{Worst} case, our method still yields the highest result (\textbf{45.9\%} at 50 IPC). 
These results further underscore the robustness of our clean-aware reweighting and trajectory-guided distillation design under severe label noise.

\begin{table}[!hb]
\vspace{-5mm}
  \renewcommand{\arraystretch}{1}
  \caption{Test accuracy (\%) on \textbf{CIFAR-100} under (a)symmetric noise (20\% and 40\%).
} 
  \centering
  \vspace{-2mm}
  \small
  \setlength{\tabcolsep}{3pt}
  \resizebox{\textwidth}{!}{
\begin{tabular}{c|l|ccc|ccc}
\hline
\multirow{2}{*}{\textbf{Noise Type}} 
& \multirow{2}{*}{\diagbox[width=7em]{\textbf{Method}}{\textbf{Noise Ratio}}}
& \multicolumn{3}{c|}{\textbf{20\%}} 
& \multicolumn{3}{c}{\textbf{40\%}} \\
&
& IPC=10 & 50 & 100
& IPC=10 & 50 & 100 \\
\hline

% ================= SYMMETRIC =================
\multirow{6}{*}{\textbf{Symmetric}}

& \textbf{Full Dataset}
& \multicolumn{3}{c|}{48.9$\pm$0.4}
& \multicolumn{3}{c}{39.9$\pm$0.2} \\

& \textbf{Subset}
& 12.4$\pm$0.4 & 21.1$\pm$0.3 & 26.2$\pm$0.3
& 9.0$\pm$0.2 & 14.6$\pm$0.3 & 17.3$\pm$0.3 \\

& \textbf{RCIG}~\citep{loo2023datasetdistillationconvexifiedimplicit}
& 41.2$\pm$0.3 & 38.4$\pm$0.3 & -
& 36.5$\pm$0.4 & 30.7$\pm$0.2 & - \\

& \textbf{DANCE}~\citep{zhang2024dancedualviewdistributionalignment}
& 45.8$\pm$0.2 & 48.0$\pm$0.3 & 47.5$\pm$0.2
& 39.7$\pm$0.2 & 42.0$\pm$0.4 & 42.6$\pm$0.3 \\

& \textbf{DATM}~\citep{guo2024losslessdatasetdistillationdifficultyaligned}
& 45.1$\pm$0.1 & 49.7$\pm$0.3 & 48.9$\pm$0.2
& 40.6$\pm$0.4 & 45.1$\pm$0.4 & 44.4$\pm$0.3 \\

& \textbf{Ours}
& \textbf{45.8$\pm$0.3} & \textbf{51.6$\pm$0.1} & \textbf{55.7$\pm$0.2}
& \textbf{43.2$\pm$0.2} & \textbf{48.4$\pm$0.4} & \textbf{52.8$\pm$0.2} \\

\hline

% ================= ASYMMETRIC =================
\multirow{6}{*}{\textbf{Asymmetric}}

& \textbf{Full Dataset}
& \multicolumn{3}{c|}{46.2$\pm$0.5}
& \multicolumn{3}{c}{33.0$\pm$0.1} \\

& \textbf{Subset}
& 12.2$\pm$0.1 & 21.9$\pm$0.5 & 26.9$\pm$0.3
& 9.5$\pm$0.4 & 15.0$\pm$0.3 & 18.9$\pm$0.3 \\

& \textbf{RCIG}~\citep{loo2023datasetdistillationconvexifiedimplicit}
& 38.9$\pm$0.4 & 37.5$\pm$0.2 & -
& 28.7$\pm$0.4 & 27.9$\pm$0.3 & - \\

& \textbf{DANCE}~\citep{zhang2024dancedualviewdistributionalignment}
& 43.8$\pm$0.3 & 46.7$\pm$0.3 & 48.2$\pm$0.4
& 32.2$\pm$0.4 & 34.2$\pm$0.2 & 35.22$\pm$0.3 \\

& \textbf{DATM}~\citep{guo2024losslessdatasetdistillationdifficultyaligned}
& 40.3$\pm$0.4 & 45.4$\pm$0.3 & 50.2$\pm$0.3
& 29.4$\pm$0.4 & 32.0$\pm$0.3 & 36.0$\pm$0.3 \\

& \textbf{Ours}
& \textbf{44.6$\pm$0.2} & \textbf{50.3$\pm$0.3} & \textbf{54.4$\pm$0.1}
& \textbf{34.7$\pm$0.1} & \textbf{36.5$\pm$0.4} & \textbf{40.7$\pm$0.4} \\

\hline
\end{tabular}
}
\vspace{-5mm}
\label{tab:vsSota CIFAR-100}
\end{table}

\begin{table}[tb]
\centering
\footnotesize
%\vspace{-2mm}
\renewcommand{\arraystretch}{0.95}
\caption{Test accuracy (\%) on \textbf{CIFAR-100N}.}
\vspace{-2mm}
\begin{tabular}{lcccccc}
\toprule
\textbf{IPC} 
& \textbf{RCIG}~\citep{loo2023datasetdistillationconvexifiedimplicit} 
& \textbf{DANCE}~\citep{zhang2024dancedualviewdistributionalignment}
& \textbf{DATM}~\citep{guo2024losslessdatasetdistillationdifficultyaligned}
& Subset
& Full Dataset
& \textbf{Ours} \\
\midrule
10 
& 37.0$\pm$0.3
& 42.0$\pm$0.3 
& 39.9$\pm$0.5
& 11.2$\pm$0.2
& \textbf{44.4$\pm$0.3}
& 41.0$\pm$0.3 \\
50
& 35.3$\pm$0.2 
& 43.6$\pm$0.3 
& 43.9$\pm$0.2 
& 20.3$\pm$0.4
& 44.4$\pm$0.3
& \textbf{45.9$\pm$0.1} \\
\bottomrule
\end{tabular}
\label{tab:vsSota CIFAR-100N}
\vspace{-3mm}
\end{table}

% On real-world noisy data, such as \textbf{CIFAR-N} in Table~\ref{tab:vsSota CIFAR-10N}, we observe that as the size of the synthetic dataset increases, most methods tend to overfit noisy labels: their accuracy saturates or even degrades with larger IPC. 
% In contrast, trajectory-matching--based approaches (DATM and ours) exhibit markedly better robustness to noise-induced overfitting and adapt more effectively as IPC grows, suggesting a stronger ability to extract clean supervisory signals from the original dataset. 
% As the noise level increases from Aggre (9\%) to Worst (40\%), all distillation methods inevitably suffer performance degradation; nevertheless, our method consistently surpasses all competitors, with the margin widening as the synthetic dataset becomes larger. 
% Although distribution-matching approaches (e.g., DANCE) remain competitive under low-noise, small-IPC settings, their performance drops sharply with higher noise or larger IPC, revealing limited robustness to label corruption. 
% By contrast, our method preserves clear advantages even under severe noise (e.g., the \textbf{Worst} setting), outperforming the best baseline by more than 3\% from IPC~50 onward. 
% Overall, these results highlight the strong noise resilience and compression robustness of our framework, yielding the most stable performance across varying noise levels and data scales, and confirming its effectiveness for noise-robust distilled supervision.

Extending this analysis, we evaluate our method on the larger-scale \textbf{Tiny-ImageNet} dataset under symmetric noise levels of 20\% and 40\% in Table~\ref{tab:vsSota-tiny}. Even under this challenging setting, our approach consistently outperforms all baseline methods. While DATM already achieves strong performance (30.4\% and 28.2\%), our method further improves the accuracy to 30.8\% and 30.1\%, demonstrating enhanced robustness to label noise and more effective extraction of clean supervisory signals under extremely low-data regimes. These results indicate that our proposed clean-aware sample reweighting and trajectory-informed distillation not only generalize to CIFAR datasets but also remain effective on higher-resolution and more challenging datasets.

\begin{table}[tb]
\centering
%\footnotesize
%\vspace{-5mm}
\renewcommand{\arraystretch}{0.9}
\caption{\small Test accuracy (\%) on \textbf{Tiny-ImageNet} under symmetric noise (IPC=10).}
\vspace{-2mm}
\begin{tabular}{lcccccc}
\toprule
\textbf{Noise} 
& \textbf{RCIG}~\citep{loo2023datasetdistillationconvexifiedimplicit} 
& \textbf{DANCE}~\citep{zhang2024dancedualviewdistributionalignment}
& \textbf{DATM}~\citep{guo2024losslessdatasetdistillationdifficultyaligned}
& Subset
& Full Dataset
& \textbf{Ours} \\
\midrule
20\% 
& 21.5$\pm$0.2 
& 22.0$\pm$0.2 
& 30.4$\pm$0.3 
& 4.4$\pm$0.2
& 30.0$\pm$0.3
& \textbf{30.8$\pm$0.2} \\
40\% 
& 17.9$\pm$0.2 
& 17.3$\pm$0.3 
& 28.2$\pm$0.2 
& 3.0$\pm$0.1
& 22.6$\pm$0.5
& \textbf{30.1$\pm$0.4} \\
\bottomrule
\end{tabular}
\label{tab:vsSota-tiny}
\vspace{-5mm}
\end{table}

\subsection{Ablation Studies}

\noindent\textbf{Impact of Each Component.}
We conduct an ablation study on CIFAR-10 under symmetric label noise levels of 20\% and 40\%, and evaluate performance across different data condensation settings (IPC $\in \{10, 50, 500, 1000\}$). 
Starting from the DATM baseline, we progressively add our two modules: \emph{Selective Guidance Reweighting} (SGR) and \emph{Teacher-Inspired Auxiliary Targets} (TIAT). As reported in Table~\ref{tab:abl_components_resize}, incorporating SGR yields consistent gains across all configurations under both symmetric and asymmetric noise.
Under symmetric noise, the improvements are particularly notable in severe and low-IPC settings (e.g., +4.7\% at 10 IPC with 40\% noise), demonstrating the effectiveness of trajectory-level denoising when label corruption is substantial.
A similar trend can be observed under asymmetric noise, where SGR consistently improves performance across IPC scales (e.g., +2.1\% at 50 IPC with 20\% noise and +1.9\% at 1000 IPC with 40\% noise). Although the absolute gains under asymmetric noise are generally smaller than those under symmetric noise, the improvements remain stable, suggesting that trajectory-level denoising is effective not only for random corruption but also for structured, class-dependent noise patterns.
Introducing TIAT on top of SGR further boosts accuracy across both noise types. Under symmetric noise, the full model achieves substantial additional improvements (e.g., +5.7\% at 500 IPC under 40\% noise). Notably, under asymmetric noise, TIAT delivers even more pronounced relative gains in high-noise regimes (e.g., +3.7\% at 1000 IPC with 40\% noise), highlighting the benefit of uncertainty-filtered auxiliary supervision when noise exhibits systematic bias.
Overall, the full model (\textbf{DATM + SGR + TIAT}) consistently outperforms the baseline across all IPC and noise settings. SGR provides strong and stable improvements in both random and structured corruption scenarios, while TIAT tends to contribute larger relative gains under higher noise ratios and more challenging regimes. These results confirm the complementary roles of trajectory-level denoising and teacher-side auxiliary supervision, and demonstrate the robustness of our framework to varying noise intensities, corruption structures, and condensation scales.
% As reported in Table~\ref{tab:abl_components_resize}, incorporating SGR yields consistent gains across all configurations, with particularly notable improvements under severe noise and scarce data (e.g., +4.7\% at 10 IPC with 40\% noise), demonstrating the effectiveness of trajectory-level denoising. 
% Introducing TIAT on top of SGR further boosts accuracy (e.g., an additional +5.7\% at 500 IPC under 40\% noise), highlighting the benefit of uncertainty-filtered auxiliary supervision. 
% Overall, the full model (\textbf{DATM + SGR + TIAT}) consistently outperforms the baseline across all IPC and noise levels, confirming the complementary roles of SGR and TIAT and the robustness of our framework to varying noise intensities and condensation scales. 
% Notably, SGR provides strong and stable improvements in all regimes, while TIAT tends to deliver larger relative gains in low-IPC settings, indicating that trajectory-level denoising and teacher-side auxiliary targets play distinct yet synergistic roles in enhancing noise-robust dataset distillation.

\begin{table}[!b]
\centering
%\footnotesize
\vspace{-5mm}
\renewcommand{\arraystretch}{1.3}
\caption{\small Ablation on CIFAR-10 under (a)symmetric noise (20\%, 40\%) across IPCs.
}
\vspace{-2mm}
\resizebox{\linewidth}{!}{
\begin{tabular}{c|l|cccc|cccc}
\hline
\multirow{2}{*}{\textbf{Noise Type}} 
& \multirow{2}{*}{\diagbox[width=7em]{\textbf{Method}}{\textbf{Noise Ratio}}}
& \multicolumn{4}{c|}{\textbf{20\%}} 
& \multicolumn{4}{c}{\textbf{40\%}} \\
& 
& IPC=10 & 50 & 500 & 1000 
& IPC=10 & 50 & 500 & 1000 \\
\hline

% ================= SYMMETRIC =================
\multirow{3}{*}{\textbf{Symmetric}} 

& \textbf{DATM} 
& 62.7 & 71.7 & 80.0 & \multicolumn{1}{c|}{81.9} 
& 58.2 & 67.2 & 74.3 & 76.7 \\

& \textbf{+ SGR} 
& 63.8 ($\uparrow$1.1) & 72.8 ($\uparrow$1.1) & 81.2 ($\uparrow$1.2) & \multicolumn{1}{c|}{83.3 ($\uparrow$1.4)} 
& 62.9 ($\uparrow$4.7) & 71.1 ($\uparrow$3.9) & 79.4 ($\uparrow$5.1) & 81.2 ($\uparrow$4.5) \\

& \textbf{+ SGR + TIAT} 
& \textbf{64.5} ($\uparrow$1.8) & \textbf{73.5} ($\uparrow$1.8) & \textbf{81.5} ($\uparrow$1.5) & \multicolumn{1}{c|}{\textbf{83.3} ($\uparrow$1.4)} 
& \textbf{64.8} ($\uparrow$6.6) & \textbf{71.6} ($\uparrow$4.4) & \textbf{80.0} ($\uparrow$5.7) & \textbf{81.4} ($\uparrow$4.7) \\

\hline

% ================= ASYMMETRIC =================
\multirow{3}{*}{\textbf{Asymmetric}} 

& \textbf{DATM} 
& 61.9 & 71.3 & 77.7 & 80.4 
& 55.9 & 63.9 & 69.7 & 71.4 \\

& \textbf{+ SGR} 
& 63.1 ($\uparrow$1.2) & 73.4 ($\uparrow$2.1) & 78.7 ($\uparrow$1.0) & 81.4 ($\uparrow$1.0) 
& 56.7 ($\uparrow$0.8) & 65.7 ($\uparrow$1.8) & 71.5 ($\uparrow$1.8) & 73.3 ($\uparrow$1.9) \\

& \textbf{+ SGR + TIAT} 
& \textbf{63.8} ($\uparrow$1.9) & \textbf{73.6} ($\uparrow$2.3) & \textbf{79.8} ($\uparrow$2.1) & \textbf{82.1} ($\uparrow$1.7) 
& \textbf{58.7} ($\uparrow$2.8) & \textbf{67.1} ($\uparrow$3.2) & \textbf{73.2} ($\uparrow$3.5) & \textbf{75.1} ($\uparrow$3.7) \\

\hline
\end{tabular}
}
\vspace{-3mm}
\label{tab:abl_components_resize}
\end{table}

\begin{figure}[!t]
    \centering
    %\vspace{-5mm}
    \includegraphics[width=1.0\textwidth]{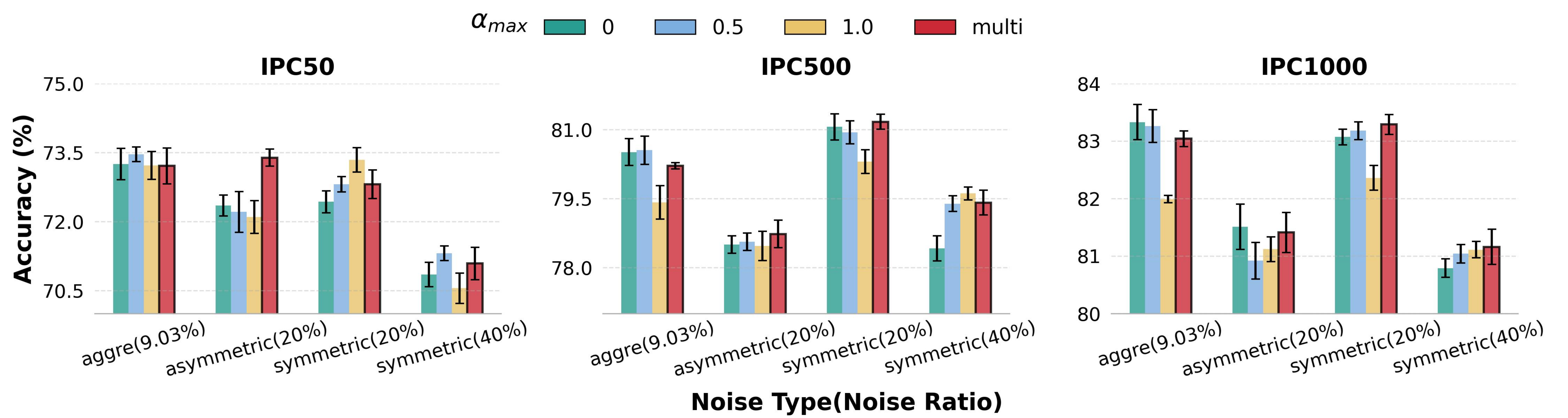}
    \vspace{-5mm}
    \caption{\small Performance comparison between \textit{Diverse Sampling} and \textit{Fixed Sampling} under various settings on \textbf{CIFAR-10}. \textbf{Baseline} refers to DATM without our method. \textbf{Diverse Sampling} and \textbf{Fixed Sampling} incorporate our \textbf{Selective Guidance Reweighting} into DATM, where Fixed Sampling uses fixed $\alpha_{\max}$ values (\textbf{0, 0.5, 1.0}) for teacher trajectory training, while Diverse Sampling \textbf{multi} follows the strategy outlined in Section\ref{section:SGR}.}
    \label{fig:difalpha}
    \vspace{-5mm}
\end{figure}

\noindent\textbf{Effectiveness of Diverse Sampling.}
To evaluate the robustness of progressive trajectory diversity, we compare \emph{Diverse Sampling}---where $P$ teacher trajectories are trained with $\alpha_{\max}$ values uniformly distributed in $[0, 1]$---against a \emph{Fixed Sampling} baseline with a constant $\alpha_{\max}$. 
As shown in Fig.~\ref{fig:difalpha}, under increasing noise levels, the Diverse Sampling strategy consistently outperforms all fixed-$\alpha$ configurations and maintains stable performance without noticeable degradation. 
In contrast, fixed-$\alpha$ trajectories exhibit varying sensitivity to the noise rate, revealing limited robustness and poor generalization. 
These results indicate that injecting diversity into teacher signal strength substantially enhances resilience to label corruption and yields more robust, adaptive supervision for dataset distillation.

\noindent\textbf{Effect of High-Confidence Subset Ratio and $\beta$ Coefficient.}
We examine how the proportion of high-confidence samples used to construct the subset $\mathcal{D}_{\text{sub}}$ and the auxiliary loss weight $\beta$ influence distillation performance under 20\% and 40\% symmetric noise on CIFAR-10 (IPC = 50). 
As shown in Table~\ref{tab:high_confidence_ratio}, our method remains robust across a broad range of subset ratios (50\%–70\%) and $\beta$ values. 
While $\beta = 0.5$ attains the highest accuracy in some cases, $\beta = 0.1$ consistently delivers stable performance across both noise levels, especially when combined with a 60\% high-confidence subset. 
In contrast, setting $\beta = 1.0$ tends to degrade performance, as it overemphasizes the auxiliary objective at the expense of trajectory alignment. 
Based on these observations, we adopt $\beta = 0.1$ and a 60\% subset ratio as default hyperparameters in all main experiments.

\begin{table}[!t]
\centering
\footnotesize
\renewcommand{\arraystretch}{1.}
\setlength{\tabcolsep}{4pt}
\caption{Ablation of $\beta$ under symmetric noise (20\%, 40\%) on CIFAR-10. 
Top two accuracies are in \textbf{bold}, and the default $\beta$ is \colorbox{cyan!10}{highlighted}.}
\vspace{-2mm}
{
\resizebox{0.95\linewidth}{!}{
\begin{tabular}{c|ccc|ccc}
\toprule
\multirow{2}{*}{\diagbox[width=6em]{\bf $\beta$}{\textbf{Subset}}}
& \multicolumn{3}{c|}{\textbf{Symmetric (20\%)}} 
& \multicolumn{3}{c}{\textbf{Symmetric (40\%)}} \\
& 50\% & 60\% & 70\% 
& 50\% & 60\% & 70\% \\
\midrule
\rowcolor{cyan!10}
$\beta$ = 0.1 & 73.3$\pm$0.11 & \textbf{73.5$\pm$0.38} & 73.3$\pm$0.4 
& 71.9$\pm$0.2 & 71.5$\pm$0.4 & \textbf{71.8$\pm$0.1} \\
$\beta$ = 0.5 & 72.5$\pm$0.29 & 73.4$\pm$0.12 & \textbf{74.0$\pm$0.3} 
& 71.7$\pm$0.3 & \textbf{72.2$\pm$0.3} & 71.4$\pm$0.2 \\
$\beta$ = 1.0 & 67.4$\pm$0.12 & 69.6$\pm$0.17 & 71.6$\pm$0.3 
& 70.0$\pm$0.3 & 71.7$\pm$0.2 & 71.5$\pm$0.1 \\
\bottomrule
\end{tabular}}
}\label{tab:high_confidence_ratio}
\vspace{-5mm}
\end{table}

\noindent\textbf{Efficiency Analysis.}
Table~\ref{tab:runtime_memory} reports runtime and peak GPU memory on CIFAR-10 using one RTX 3090. Since both DATM and our method use the same TESLA-style memory-saving implementation, SGR/TIAT introduce negligible additional peak memory; the main cost is training-time overhead from reliability estimation and auxiliary target construction.

\begin{table}[htb]
\centering
\vspace{-6mm}
\caption{\small Runtime and peak GPU memory analysis.}
\vspace{-2mm}
\scriptsize
\setlength{\tabcolsep}{3pt}
\renewcommand{\arraystretch}{1.0}
\begin{tabular}{c|c|c|c|c}
\toprule
\textbf{IPC} & \textbf{Peak} & \textbf{DATM} & \textbf{Ours} & \textbf{Overhead} \\
             & \textbf{Mem.} & \textbf{Time} & \textbf{Time} &  \\
\midrule
10   & 2.5GB  & 6.5h  & 7.0h  & 1.08$\times$ \\
50   & 6.2GB  & 17h   & 25h   & 1.47$\times$ \\
500  & 10.4GB & 24h   & 36h   & 1.50$\times$ \\
1000 & 11.6GB & 60h   & 70h   & 1.17$\times$ \\
\bottomrule
\end{tabular}
\vspace{-4mm}
\label{tab:runtime_memory}
\end{table}

\section{Conclusion}
We investigate the underexplored problem of dataset distillation under noisy-label settings and identify two key challenges: overfitting to label noise and limited capacity of the synthetic set to retain clean signals. To address these, we introduce Selective Guidance Reweighting and Teacher-Inspired Auxiliary Targets. Experiments on benchmark datasets validate the robustness and effectiveness of our approach, paving the way for future research in noise-resilient dataset distillation.

% In this work, we conduct a focused investigation into the problem of dataset distillation under noisy-label settings, a novel and underexplored research direction. Through systematic analysis, we identify two key challenges that limit the effectiveness of existing distillation approaches in noisy environments: (1) the synthetic set overfit the noise distribution present in the original data, thereby compromising distillation quality; and (2) the inherent capacity bottleneck of the synthetic set, which restricts its ability to capture clean and informative signals. To address these challenges, we propose two targeted solutions: Selective Guidance Reweighting and Teacher-Inspired Auxiliary Targets. Extensive experiments on widely used dataset distillation and noisy-label learning benchmarks demonstrate the robustness and superior performance of our method, laying a solid foundation for future research in noise-resilient dataset distillation.

\section*{Acknowledgment}
This work has been supported in part by the National Natural Science Foundation of China (Grant No. 62472139, 62502142, 62502144), the Natural Science Foundation of Anhui Province (Grant No. 2508085QF226), the Open Project Program of the State Key Laboratory of CAD\&CG (Grant No. A2403), Zhejiang University. The computation is completed on the HPC Platform of Hefei University of Technology.

\bibliographystyle{splncs04}
\bibliography{main}

% {
%     \small
%     \bibliographystyle{splncs04}
%     \bibliography{main}
% }

\end{document}